# THEME-AWARE GENERATION MODEL FOR CHINESE LYRICS

*Jie Wang[1,2], Xinyan Zhao[3]*

[1]*Department of Electrical Engineering, The Hong Kong Polytechnic University*
[2]*Tencent Group*
[3]*School of Data Science, University of Science and Technology of China*

*jayjay.wang@connect.polyu.hk, sa516458@mail.ustc.edu.cn*

**ABSTRACT**

With rapid development of neural networks, deep-learning has been extended to various natural language generation fields, such as machine translation, dialogue generation and even literature creation. In this paper, we propose a theme-aware language generation model for Chinese music lyrics, which improves the theme-connectivity and coherence of generated paragraphs greatly. A multi-channel sequence-to-sequence (seq2seq) model encodes themes and previous sentences as global and local contextual information. Moreover, attention mechanism is incorporated for sequence decoding, enabling to fuse context into predicted next texts. To prepare appropriate train corpus, LDA (Latent Dirichlet Allocation) is applied for theme extraction. Generated lyrics is grammatically correct and semantically coherent with selected themes, which offers a valuable modelling method in other fields including multi-turn chatbots, long paragraph generation and etc.

## 1. INTRODUCTION

In recent years, natural language generation (NLG) has been recognized as an important but challenging task in the field of Artificial Intelligence (AI). There are many reports about poetry generation (Wang et al., 2016; Yan, 2016; Yi et al., 2016), response generation (Shang et al., 2015; Wang et al., 2018), machine translation (Tu et al., 2016) and so on. As an effective NLG model, the encoder-decoder framework (Cho et al., 2014) has been extensively exploited which can generate words one by one. However, few studies have been conducted on song lyrics generation. As a special literature, song lyrics owns both literary and musical attributes. In general, lyrics are composed of main and chorus sections. A common format of lyrics is one main section followed by one chorus or two main sections followed by one chorus. Compared with classical Chinese poetry, paragraph structures of lyrics are more free and longer, which result in semantically coherent generation more difficult.

In this paper, we proposed a theme-aware generation model which can compose long-paragraph and semantically-coherent Chinese lyrics. Similar to poetry generation, an attention-based Seq2Seq model has been used to generate lyrics line by line. However, semantic drift may occur when generated sentences accumulate. To overcome this problem, a multi-channel encoding structure has been incorporated into the generation model, which encodes theme keywords and previous sentences as global and local context vectors respectively. Afterwards, a sentence decoder with attention mechanism will output next lyrics line by line.

The key contributions of this paper are summarized as follows:

- A theme-aware seq2seq model has been proposed, with a multi-channel encoder encoding thematic keywords and local context to control the overall semantics of generated paragraphs. To the best of our knowledge, this might be the first work of thematic controlling on lyrics generation.
- To enable self-learning rhyming structures of lyrics for the model, arrangement of sentence pairs in train corpus has been optimized.

The rest of this paper is organized as follows. Section 2 discusses related work, including similar studies such as Chinese poetry and dialogue generation systems. Section 3 describes train corpus preparation and theme-aware seq2eq model. We present experimental results in Section 4. Section 5 concludes the paper.

## 2. RELATED WORK

As a special kind of NLG application, lyrics generation is similar to dialogue and Chinese poetry generation. Among them, dialogue generation have been advanced considerably due to the fast development of deep learning and available large corpus. Encoder-decoder based framework was widely applied to solve single-turn dialogue tasks [1], [2]. Particularly, some novel models were proposed to consider conversational history in multi-turn dialogue systems.

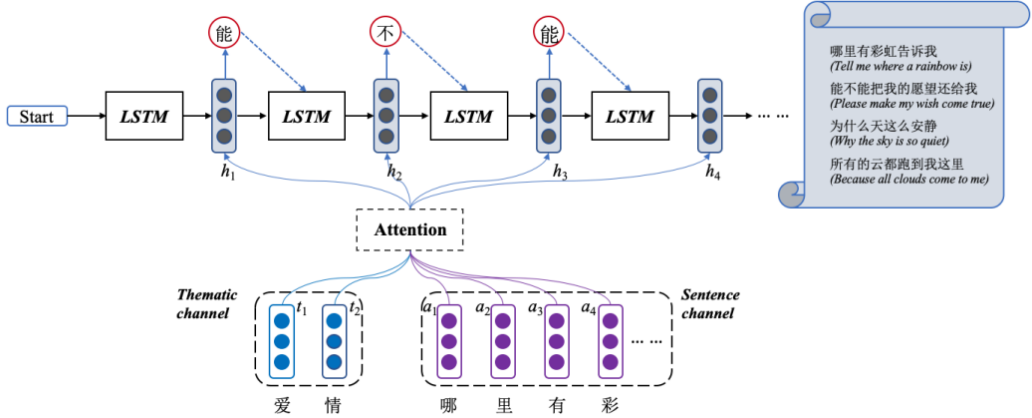

**Figure 1**: Schematic illustration of the thematic-aware generation based on a multi-channel Seq2Seq framework.

Sordoni et al. encoded the contextual history into fixed-length vectors and fed into a recurrent neural network (RNN) based language model for response prediction [3]. HRED use one RNN at the word-level and another at the utterance level to model the hierarchical structure of contexts [4]. To deepen and widen topics in conversation, Wang et al. presented a deeper encoding scheme with three channels, namely global, wide, and deep ones [5]. However, an existing Gordian knot of this area is keeping semantic and logical consistency during generation.

On the other hand, similar studies of Chinese poetry generation have also been conducted. For instance, Zhou et al. used a genetic algorithm for poetry generation with tonal codings and state search [6]. Jiang and Zhou proposed a statistical machine translation (SMT) model to predict next sentences based on previous ones [7]. Yi et al. considered Chinese poetry generation as a sequence-to-sequence learning problem, and built an encoder-decoder framework to generate quatrains [8]. Moreover, Wang et al. demonstrated a two-stage poetry generation scheme which consists of sub-topic planning and poem line writing [9]. Furthermore, Yan incorporated a polishing schema into a generation model to imitate human writing process [10].

To improve the overall thematic connectivity and logic coherence of generated long-paragraph texts, we proposed a thematic aware and multi-channel Seq2Seq model for lyrics generation. Moreover, this model is able to self-learn rhyme structures of lyrics automatically.

### 3. MODEL

To enhance semantic consistency of generation, we propose a multi-channel encoder to encode both topic keywords and previous sentences in the Seq2Seq scheme, as illustrated in **Figure 1**. Before the RNN decoder, a multi-channel encoder consists of thematic and lyrics sentence encoders. In the train phase, topic keywords can be extracted from lyrics sentences with an optimized LDA model. In the inference phase, topic keywords and preceding sentences are respectively fed into these two encoders and converted into fixed-length hidden vectors. Moreover, an attention mechanism is applied to the RNN decoder to automatically focus on different words of input contexts. In this section, each part of this generation model will be discussed at detail.

**3.1 Extraction of Thematic Keywords**

Owing to its unsupervised essence, LDA is a popular method for text topic analysis, which is based on a hypothesis that each item of one collection can be modeled as a finite mixture of hidden random variables [11]. A parameter $k$, i.e. topic number, needs to be pre-determined. In this paper, the parameter $k$ is set as 6. All of lyrics corpus is used to train this LDA model before stop words were omitted. The first-ranking phrases of each topic were selected as corresponding thematic keywords, which are shown in **Table 1**.

| Original song lyrics | Preceding lyric sentence | Next lyric sentence |
|---|---|---|
| 哪里有彩虹告诉我 *(Tell me where a rainbow is)* | 哪里有彩虹告诉我 *(Tell me where a rainbow is)* | 能不能把我的愿望还给我 *(Please make my wish come true)* |
| 能不能把我的愿望还给我 *(Please make my wish come true)* | 能不能把我的愿望还给我 *(Please make my wish come true)* | 为什么天这么安静 *(Why the sky is so quiet)* |
| 为什么天这么安静 *(Why the sky is so quiet)* | 为什么天这么安静 *(Why the sky is so quiet)* | 所有的云都跑到我这里 *(Because all clouds come to me)* |

**Table 1**. Examples of lyric sentence pairs.

| Theme ID | Keywords |
|---|---|
| 0 | 等待 (Waiting) |
| 1 | 梦想 (Dream) |
| 2 | 爱情 (Love) |
| 3 | 喜欢 (Fondness) |
| 4 | 家乡 (Hometown) |
| 5 | 寂寞 (Loneliness) |

**Table 2**. The first-rank keywords of each theme.

## 3.2 Lyrics Generation

In a probabilistic view, the optimization of a Seq2Seq model can be considered as maximizing the likelihood of observing the output (target) sequence given an input (source) one. As shown in **Figure 2**, an RNN encoder-decoder network is used to train the generation model based on the preprocessed (keywords, sequence pair) corpus. That is to say, this trained model can generate a new lyrics line given input thematic keywords and a previous line, i.e. line by line.

The multi-channel encoder module consists of two bi-directional long short-term memory network (Bi-LSTM) cells, namely topic and sentence encoders [12]. Among them, the topic encoder encodes global thematic information while the sentence one encodes local contextual information. For instance, given preceding line *src* with length of $L_{src}$, i.e. *src* = $\{src_1, src_2,..., src_{L_{src}}\}$, next line *trg* with length of $L_{trg}$ should be predicted. As shown in **Figure 2**, the source sentence *src* is converted into a vector of hidden states $\{h_1, h_2, ..., h_{L_{src}+1}\}$ with dimension of $L_{src}$. Similar to the sentence encoder, topic encoder take in thematic keywords $\{m_1, m_2\}$. Note that all topic keywords compromise two Chinese characters.

As the decoder, another one-layer LSTM cell take in the last hidden state from the encoder as initialization. Hidden state vectors from two encoders are concatenated as $\{h_1, h_2, ..., h_{L_{src}+1}; m_1, m_2\}$, as the input of the attention layer. In generation step $t$, output $y_t$ of the decoder is determined by the hidden state $s_t$ of the $t$ step, context vector $c_t$ and last-step output $y_{t-1}$, as expressed as follows:
$$y_t = argmax_y P(y|s_t, c_t, y_{t-1}).$$
After each step, $s_t$ is updated by
$$s_t = f(s_{t-1}, c_{t-1}, y_{t-1}),$$
and $f(.)$ is an update function of LSTM cell. The context vector $c_t$ of the attention layer is recomputed at each step:
$$c_t = \sum_{j=1}^{L_{src}+1} \alpha_{t,j} h_j + \alpha_{t, L_{src}+2} m_1 + \alpha_{t, L_{src}+3} m_2.$$
The weight $\alpha_{t,j}$ is computed by
$$\alpha_{t,j} = \frac{\exp(e_{t,j})}{\sum_{k=0}^{L_{src}+2} \exp(e_{t,k})}.$$

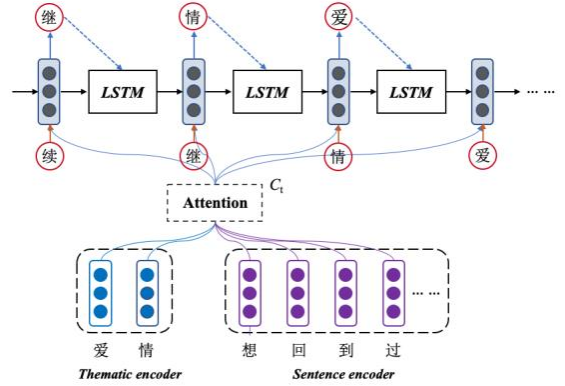

**Figure 2.** Illustration of the MC-Seq2Seq model with attention mechanism. Input texts consist of "想回到过去" (I want to go back to the past) and theme of "爱情" (love). Target sentence of "让爱情继续" (Keep the love going on) is reversed for rhythm controlling.

Where $e_{t,j}$ is an attention scoring function and can be an additive model:
$$e_{t,j} = v_a^T \tanh(W s_{t-1} + U h_j),$$
or dot-product model:
$$e_{t,j} = s_{t-1} W h_j.$$
Where $W$, $v_a^T$, $U$ are trainable parameters. Although additive and dot-product attention are similar in terms of complexity, the dot-product one is often faster and more effective because of more effective matrix operation [13].

## 4. EXPERIMENTS

### 4.1 Dataset

Around lyrics of 140 thousand songs has been collected after duplication omitting. As depicted in **Table 2**, each lyrics sample was preprocessed into the form of sequence pair and then with data cleaning. Qualified lyrics sequence pairs are selected according to the following rules: (1) The lyrics full of stop words should be omitted. For example, lyrics like "啦啦啦" (La La La) were deleted; (2) lyrics sentences with 3~18 Chinese characters should be retained; (3) repeated sentences have been removed. At last, three million of sequence pairs has been gathered covering six themes.

### 4.2 Experimental Settings

The model training objective is minimizing the cross-entropy loss between predicted sentence and ground truth. The model parameters were updated by the back-propagation method. In this paper, word embedding has been pre-trained using word2vec based on the whole lyrics corpus [14]. Note that Chinese characters with occurrence times less than 5 are omitted as unknown words. As a result, the vocabulary size is approximately 5,000. Both encoders and decoder share the same vocabulary. The word embedding dimension size is 512. The hidden layer sizes of two encoders and the decoder all are 1024. ReLU function is utilized as the activation function in neural networks. The initial learning rate is set as 0.1. Parameters of our model were randomly initialized over a uniform distribution with support [-0.1,0.1]. The model was trained with AdaDelta algorithm [15], and mini-batch was set to be 320.

| Model | Score | | | | | |
|---|---|---|---|---|---|---|
| | Topic-Integrity | Topic-Relevance | Fluency | Coherence | Lyricism | Average |
| Seq2Seq (Baseline) | 2.38 | 2.76 | 3.99 | 3.15 | 3.44 | **3.14** |
| MC-Seq2Seq | 3.18 | 3.92 | 4.15 | 3.36 | 3.96 | **3.71** |

Table 3. Averaged ratings for Chinese lyrics generation between Seq2Seq (Baseline) and MC-Seq2Seq (our model).

### 4.3 Result

As shown in **Figure 3**, convergence state has been achieved after 150 epochs. Moreover, several generation examples are shown in **Table 3** to verify the effectiveness of the thematic-aware model. In fact, because most of train songs are related to love theme, generated samples will be gradually tending to be love-related without explicit interruption. As verified by generated samples, our model explicitly takes into account global thematic contexts. With this thematic guidance, our model generates more meaningful results instead of general paragraphs. Furthermore, it is noteworthy that this model can self-learn rhythm structures even if no rules were explicitly applied. If the order of input sentences were reversed, rhyme patterns could be more dominant in generated paragraphs.

Since lyrics generation seems like a kind of literature creation, Bilingual Evaluation Understudy (BLEU) or other evaluation metrics are not suitable for our case. Here, human evaluation is conducted by 5 Chinese experts and also pop music fans. Similar to reference, 500 randomly selected lyrics samples are rated by these experts with a 1-5 scale on five dimensions: "Topic-Integrity", "Topical-Relevance", "Coherence", "Fluency" and "Lyricism". The score of each dimension ranges from 1 to 5 with higher score the better. And final score is achieved with the average of rating scores, as shown in **Table 3**.

### 5. CONCLUSION AND FUTURE WORK

**Table 4:** Generation samples under the six themes.

Theme: 爱情 (*Love*)
哪里有彩虹告诉我 (*Tell me where a rainbow is*)
明天是什么颜色 (*What color at tomorrow*)
你不属于我 (*Your heart does not belong to me*)
不是我 (*Not for me*)
你会不会软弱 (*Will your heart be weak*)
不要再沉默 (*Don't be silent again*)

Theme: 家乡 (*Hometown*)
哪里有彩虹告诉我 (*Tell me where a rainbow is*)
是心底里的形状 (*Its shape appears in my mind*)
给我光辉曙光 (*Give me a glowing light*)
飞到你身旁 (*I will fly to you*)
亲爱的姑娘 (*My dearest girl*)
我的家乡 (*Fly to my hometown with me*)

Theme: 梦想 (*Dream*)
哪里有彩虹告诉我 (*Tell me where a rainbow is*)
是心底的形状 (*Its shape appears in my mind*)
给我光辉曙光 (*Give me a glowing light*)
浪失去一点希望 (*Lost hope along with waves*)
没有阳光 (*Without sunshine*)
没法能阻挡 (*No way to stop me following my dream*)

Theme: 等待 (*Waiting*)
哪里有彩虹告诉我 (*Tell me where a rainbow is*)
是心底里的形状 (*Its shape appears in my mind*)
给我绘上曙光 (*Drawing a glowing light for me*)
新的希望 (*New hope*)
就我们在一起 (*Let's stay together*)
快乐的在一起 (*To be happy*)

Theme: 喜欢 (*Fondness*)
哪里有彩虹告诉我 (*Tell me where a rainbow is*)
是心底里的形状 (*Its shape appears in my mind*)
给我绘上曙光 (*Drawing a glowing light for me*)
新的希望 (*New hope*)
在一起 (*To be together*)
我们在一起 (*We love each other*)

Theme: 寂寞 (*Loneliness*)
哪里有彩虹告诉我 (*Tell me where a rainbow is*)
明天是什么颜色 (*What color at tomorrow*)
世界只剩我一个 (*If I were the last one on earth*)
相濡以沫 (*Let's help each other*)
别再相互折磨 (*Don't torture each other*)
夜深人静的时候 (*Even in the deep night*)

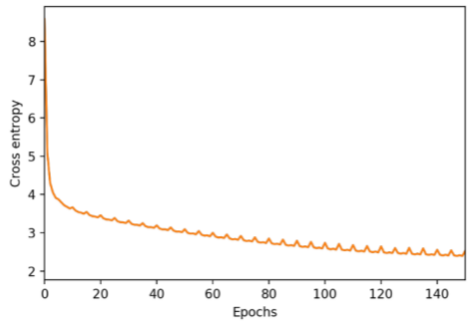

**Figure 3**. Evolution of training loss against epochs.

In this paper, we propose a thematic-aware Seq2Seq model, which can generate new high-quality lyrics with perfect topic connectivity. A multi-channel encoder has been constructed by fusing two different level contextual information. Fluent generated lyrics samples also demonstrate the efficacy of our framework. Moreover, this generation model also self-learns the rhyme patterns. With a detailed analysis, we find stronger rhyme information close to the decoder can be achieved by reversing the input sentence order.

As future work, we will shed light on how to incorporate paragraph structure information into the generation model. Moreover, logical and semantic consistency throughout long paragraphs still need more efforts.